\documentclass[conference]{IEEEtran}
\IEEEoverridecommandlockouts
\usepackage{cite}
\usepackage{amsmath,amssymb,amsfonts}
\usepackage{algorithmic}
\usepackage{graphicx}
\usepackage{textcomp}
\usepackage{xcolor}
\usepackage{booktabs}
\usepackage[hidelinks]{hyperref}
\usepackage{multirow}
\usepackage{tcolorbox}
\usepackage{tabularx}
\usepackage{subcaption}
\usepackage{soul}
\usepackage{float}
\usepackage{newtxmath} 
\usepackage{threeparttable}
\newcommand{\res}[2]{$#1_{#2}$} 
\newcommand{\bestres}[2]{$\mathbf{#1}_{#2}$} 
\newcommand{\secres}[2]{$\underline{#1}_{#2}$}

\def\BibTeX{{\rm B\kern-.05em{\sc i\kern-.025em b}\kern-.08em
    T\kern-.1667em\lower.7ex\hbox{E}\kern-.125emX}}
\begin{document}

\title{LLM-Powered Personalized Glycemic Assessment in Type 2 Diabetes with Wearable Sensor Data \\
}

% \author{

% \IEEEauthorblockN{Yifan Gao}
% \IEEEauthorblockA{
% \textit{Department of Information Systems} \\
% \textit{and Cybersecurity} \\
% \textit{The University of Texas at San Antonio}\\
% San Antonio, USA\\
% yifan.gao@utsa.edu
% }

% \and

% \IEEEauthorblockN{Yanmin Gong}
% \IEEEauthorblockA{
% \textit{School of Engineering Medicine} \\ 
% % \textit{Department of Computer Science}\\
% \textit{Texas A\&M University}\\
% Houston, USA\\
% yanmin.gong@tamu.edu
% }

% \and

% \IEEEauthorblockN{Yun Shi}
% \IEEEauthorblockA{
% \textit{Long School of Medicine}\\
% \textit{The University of Texas at San Antonio}\\
% San Antonio, USA\\
% shiy@uthscsa.edu
% }

% \and

% \IEEEauthorblockN{Yuanxiong Guo}
% \IEEEauthorblockA{
% \textit{Department of Information Systems} \\
% \textit{and Cybersecurity}\\
% \textit{The University of Texas at San Antonio}\\
% San Antonio, USA\\
% yuanxiong.guo@utsa.edu
% }

% }

\author{%
\begin{minipage}{\linewidth}
\centering
\begin{minipage}[t]{0.46\linewidth}\centering
Yifan Gao\\
\textit{Department of Information Systems}\\
\textit{and Cybersecurity}\\
\textit{The University of Texas at San Antonio}\\
San Antonio, USA\\
yifan.gao@utsa.edu
\end{minipage}\hfill
\begin{minipage}[t]{0.46\linewidth}\centering
Yanmin Gong\\
\textit{School of Engineering Medicine}\\
\textit{Texas A\&M University}\\
Houston, USA\\
yanmin.gong@tamu.edu
\end{minipage}\\[1.5em]
\begin{minipage}[t]{0.46\linewidth}\centering
Yun Shi\\
% \textit{Long School of Medicine}\\
\textit{Department of Family}\\
\textit{and Community Medicine}\\
\textit{The University of Texas at San Antonio}\\
San Antonio, USA\\
shiy@uthscsa.edu
\end{minipage}\hfill
\begin{minipage}[t]{0.46\linewidth}\centering
Yuanxiong Guo\\
\textit{Department of Information Systems}\\
\textit{and Cybersecurity}\\
\textit{The University of Texas at San Antonio}\\
San Antonio, USA\\
yuanxiong.guo@utsa.edu
\end{minipage}
\end{minipage}%
}

\maketitle
% normalsize(10pt) > small(9pt) > footnotesize(8pt)

\begin{abstract}
Type 2 Diabetes (T2D) poses an increasing global health threat, demanding effective glycemic assessment to support personalized and improved diabetes care. Wearable sensors such as continuous glucose monitors (CGM) and fitness trackers offer many valuable insights for glycemic assessment. However, effectively analyzing these data requires integration with essential individual-level context. Existing methods are often based on traditional machine learning (ML) and rely primarily on historical blood glucose measurements and overlook personalized information, which limits their performance across diverse diabetes populations.
Recent advances in large language models (LLMs) have demonstrated their ability to integrate diverse data modalities while modeling sequential dependencies, motivating the exploration of their potential for personalized glycemic assessment. 

In this paper, we propose GlyLLM, an LLM-powered framework for modeling CGM-based glycemic dynamics through the integration of wearable sensor data and structured metadata. GlyLLM can leverage the extensive prior knowledge of pre-trained LLMs and achieve sensor-text semantic abstraction at decision time. 
Experiments on two related tasks on the AI-READI dataset demonstrate that our model outperforms traditional ML methods by an average of 13.66\% in Root Mean Squared Error (RMSE)
for glucose forecasting and 13.08\% in Area Under the
Receiver Operating Characteristic (AUROC)
for diabetes categorization. 
Additionally, our ablation study shows that diabetes surveys and biometric tests are more critical than other health information 
for glycemic assessment. Our work presents a promising step toward harnessing the power of LLMs to advance personalized glycemic assessment in T2D care.
\end{abstract}


\section{Introduction}

% \yifan{Reference paper: \textbf{LLM-powered personalized glucose prediction in type 1 diabetes}}
% \yifan{First paragraph: introduce Type 2 Diabetes disease and its public health threat.}
Type 2 Diabetes (T2D) is a growing chronic metabolic disorder characterized by hyperglycemia due to reduced insulin secretion and/or insulin resistance. As reported by the International Diabetes Federation (IDF), the number of people with diabetes was estimated at 589 million in 2024, with T2D accounting for over 90\% \cite{Huang2025}. Therefore, T2D has been a significant public health threat that requires much attention and intervention. 
Poorly controlled diabetes increases the risk of cardiovascular complications and serves as a significant risk factor for other diseases, including chronic kidney disease, various cancers, and mental health disorders \cite{Tomic2022, Parker2023}. In addition, it places  substantial financial pressure on both individuals and the global healthcare system, with total expenditure on diagnosed diabetes in the United States reaching 412.9 billion USD in 2022, representing a 35\% increase over the last decade \cite{Parker2023}. As the impact of diabetes continues to grow, comprehensive glycemic assessment is essential to manage the disease and improve patient care.
% Effective glycemic management is crucial for individuals with Type 2 Diabetes Mellitus (T2DM) prevent both immediate risks, such as hypoglycemia and hyperglycemia. 

% \yifan{Second paragraph: introduce why analyzing sensor data and considering personalized information are clinical and reasonable for glycemic assessment. } \guo{(This paragraph should emphasize the importance of wearable sensors, particularly CGM, in diabetes management. )} \yun{I agree with Dr. Guo, the importance of CGM in diabetes care, touching on the different populations, such as T1DM, insulin-treated T2DM, non-insulin-treated DM, and the proposed expansion of prediabetes.} 
% \yifan{{CGM benefits/LLM benefits}}
% It is well-known that self-monitoring of blood glucose and other vital signs of metabolic health is crucial to support diabetes care \cite{Kaufman2023}. 
Wearable devices such as continuous glucose monitors (CGMs) and physical activity trackers are now widely used to monitor 
% \hl{patients'} \guo{(Later you just said extended to other population. So what types of patients are you referring to here?)} 
insulin-treated diabetes patients' real-time glucose fluctuations and related vital sign trends \cite{rodriguez2021mobile, Wang2023}. 
% Recently, interest in CGM has extended into populations \hl{with and at-risk of non-insulin treated diabetes} \guo{(what does this mean? grammatical error.)} 
Recently, interest in CGM has extended into non-insulin-treated and prediabetes populations, further supported by the approval of over-the-counter CGM devices in the United States.
% \cite{Shah2025}. 
While insulin-treated patients often rely on CGM for tight glycemic control \cite{Healey2025}, non-insulin diabetes and prediabetes populations increasingly use CGM to understand glycemic variability and long-term metabolic patterns \cite{klonoff2025continuous}. The resulting data provide rich physiological information, forming a basis for glycemic assessment.

% \yifan{Fourth paragraph: Recent work for analyzing wearable sensing data in diabetes and they overlook the personalized metadata.}
% \yifan{logic}
Recently, various traditional machine learning (ML) methods have been proposed to analyze wearable sensor data for diabetes 
\cite{Shuvo2023, Metwally2024, Zhu20251}.
However, most existing methods primarily rely on historical episodic blood glucose measurements while overlooking individual-level static metadata. 
% Beyond \hl{these task-specific limitations} \guo{(What do you mean by these?)}, \hl{such}\guo{(what do you mean by such?)} dataset-specific models suffer from limited performance due to their reliance on small labeled datasets and the requirement for expensive retraining for each new task. \guo{(This is an important part and need to be very clear.)}
Beyond this limitation, such ML-based models suffer from limited performance due to their reliance on small labeled datasets and the requirement for expensive retraining for specific tasks.

LLMs have demonstrated significant potential in health applications, as their massive pre-training allows for generalization across a wide range of downstream tasks with minimal task-specific data. However, when applying LLMs, current standard prompting-based approaches 
\cite{Alavi2024, Healey2024, cardei2025dm, Zhu2025} remain ineffective for integrating static metadata with long-horizon sensor data for diabetes-related tasks. Static metadata and wearable sensor data are distinct modalities that provide complementary information, and effective techniques capable of integrating them are required.

% \yifan{Fifth paragraph: Summarize our work contribution to explore the potential of LLM for personalized glycemic assessment.} \guo{(You will need to mention about the tasks you are focusing on in this paper, and why you study them.)} 
To overcome these limitations, we propose GlyLLM, an LLM-powered framework that leverages the capabilities of pre-trained LLMs to jointly interpret personalized static metadata and wearable sensor data collected from patients with T2D, as illustrated in Fig.~\ref{fig:Model Architecture}. 
To enable effective wearable data semantic extraction, we generate sequences of wearable data representations via a pre-trained vision transformer (ViT) encoder and concatenate them with personalized static metadata representations as unified inputs for the backbone LLM in the framework. 
% \hl{In terms of complementary glycemic assessment} \guo{(what do you mean want to say here by in terms of?  What is so called complemenatry?)} to reflect both short-term and long-term clinical decision needs in T2D, we first consider glucose level forecasting to evaluate the model's capability to detect the glucose trend. \hl{In the second stage} \guo{(two stages means they are sequential decisions. Here we do not have. Moreover, we never mention about the first stage before. Frist does not mean first stage.)}, diabetes categorization can help identify long-term metabolic status.
For comprehensive glycemic assessment to reflect both short-term and long-term clinical decision needs in T2D, we first consider glucose level forecasting to evaluate the model's capability to detect the glucose trend. Additionally, diabetes categorization is evaluated to help identify long-term metabolic status.
The key contributions of this paper are summarized as follows:
\begin{itemize}
% [itemsep=0pt, parsep=0pt, topsep=2pt, partopsep=2pt]
    \item We propose GlyLLM, an LLM-powered framework that integrates personalized static metadata and wearable sensor data, enabling effective adaptation of LLMs to a wide range of diabetes-related tasks. 

    % \item \guo{Add a pargraph about the technical novelty of your GlyLLM approach.}
    \item GlyLLM employs a specialized sensor encoder (ViT) designed to model both local and long-range temporal dependencies across multiple time scales. Additionally, we design a static metadata template including diabetes-related factors to facilitate personalized glycemic assessment.
    
    \item We conduct extensive experiments on the real-world AI-READI v2.0.0 dataset. The results across two distinct clinical tasks demonstrate that our approach achieves superior or competitive performance compared to state-of-the-art baselines. 
    
    \item We perform comprehensive ablation studies to quantify the contributions of personal static metadata components and sensor data within the framework, providing insights into the impact of each data type on glycemic assessment.
\end{itemize}

\section{Related Work}
\vspace{-5pt}
% \subsection{ML for CGM Data \guo{(ML for CGM Data?)}}
\subsection{ML for CGM Data}
% \yifan{Reference paper: \textbf{LLM-powered personalized glucose prediction in type 1 diabetes}}
% \yifan{This section focus on traditional machine learning and deep learning work for diabetes management using CGM glucose data.}
CGM technology has transformed personal diabetes management, enabling continuous tracking of real-time glucose trends \cite{rodriguez2021mobile}. To advance computational analysis, CGM data can be interpreted in various forms, ranging from statistical metrics (e.g., mean, standard deviation, and time-in-range) \cite{Healey2025} to functional data analysis \cite{klonoff2025continuous}. Based on these data representations, many ML approaches have been developed to analyze glucose dynamics in different tasks \cite{Metwally2024, Shuvo2023,Zhu20251}. 
For instance, 
% \guo{(use first author's last name and et al. such as ``Guo et al.'' --> Change throughout the paper. Do not use [xx] directly to start a sentence.)}
Shuvo et al. \cite{Shuvo2023} design a stacked long short-term memory (LSTM) for predicting blood glucose concentration based on historical CGM measurements. 
Metwally et al. \cite{Metwally2024} optimize a support vector machine (SVM) classifier to identify metabolic subphenotypes using CGM data derived from oral glucose tolerance tests (OGTT).
GPFormer \cite{Zhu20251} applies a customized sparse multi-head self-attention (MHSA) mechanism to better predict adverse glycemic events.
Despite the progress, existing methods for integrating wearable sensor data with static metadata remain insufficient, which limits their practical applicability due to the complex glycemic dynamics associated with inter-individual variability. 
Current approaches often overlook key aspects of modality heterogeneity, including the explicit modeling of temporal dependencies across multivariate wearable signals. As a result, they fail to fully interpret the rich and complementary information available across modalities.

\subsection{LLMs for Health}

% \yifan{Reference paper: \textbf{Based on my previous work submission}}
% \yifan{This section focus on why here we consider to explore the potential of LLM for personalized glycemic assessment.}
% \yifan{First paragraph: introduce the utilization of LLM is promising in many healthcare domains. } 
LLMs have recently gained significant attention in healthcare applications \cite{xu2024mental, kim2024health}. For example, Mental-LLM \cite{xu2024mental} evaluates multiple LLMs on a range of mental health prediction tasks using online text data, and Health-LLM \cite{kim2024health} investigates the effectiveness of LLMs in addressing consumer health questions through advanced prompt engineering techniques. These studies highlight the promise of LLMs, which benefit from large-scale pretraining on massive datasets, enabling a wide spectrum of downstream clinical health tasks.   

% \yifan{LLMs process CGM data}
% \yifan{Second paragraph: introduce emerging line of work for repurposing LLM to analysis wearable sensor data with personalized information. LLM for CGM data and health-LLM in CGM.} 
A few of the most recent studies have re-purposed LLMs to process CGM data within a unified framework \cite{Alavi2024, Healey2024, cardei2025dm}. %
%
%
%
%\hl{CGM data can be input as raw time-series text, whereas textual information is used directly. Typical strategies involve combining template-based prompts, instructing LLMs to jointly model heterogeneous signals} \guo{(Have no clue on what you want to say here.)} \cite{Alavi2024, Healey2024, cardei2025dm}. 
% \yifan{summarize, not only cite 15-16;}
Most of them view CGM data as raw time-series text and use template-based prompts to instruct LLMs to jointly model heterogeneous signals. For instance, 
% \guo{last name et al.}\cite{Alavi2024} \hl{demonstrated} \guo{(what tense shall we use? use chatgpt to polish)} 
Alavi et al. \cite{Alavi2024} demonstrate
how diabetes wearable data can be represented as text and directly processed by LLMs for health prediction tasks. LLM-CGM \cite{Healey2024} and DM-Bench \cite{cardei2025dm} develop an LLM-CGM question-answering (QA) benchmark that utilizes in-context learning to perform reasoning over CGM data understanding. Although these studies have shown promise for understanding short-duration CGM data, long-duration trends and diabetes status still require specialized domain knowledge and explicit modeling of glucose dynamics to be analyzed. %
%
%Therefore, representing sensor data as text has the following limitations. First, LLMs' performance decreases with longer CGM data understanding. Second, effective prompting is challenging and requires strong domain knowledge. 
% Although \cite{Li2025 ,Zhu2025} explored CGM modeling by wearable sensor encoders with LLMs, the impact of personalized information matters more is yet to be explored.
Our work offers an alternative perspective to repurpose LLMs for integration, which effectively extracts semantic abstraction of long-horizon CGM data dynamics into clinically meaningful outcomes. 
% \vspace{-6pt}

\section{Methodology}
In this section, we introduce the model architecture and implementation details of the proposed GlyLLM framework. 

\begin{figure*}[htbp]
    \centering
    \includegraphics[width=1.0\linewidth]{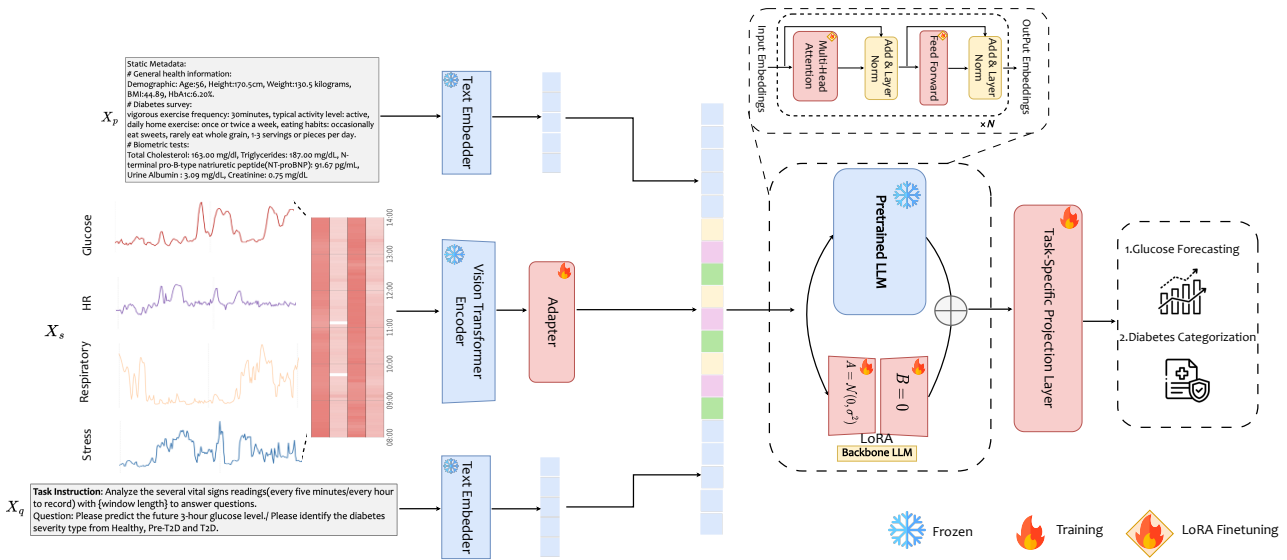}
    \caption{\textbf{GlyLLM Model Architecture.} Text embeddings from static metadata and sensor data embeddings, along with text embeddings from task instruction prompts, are structured as sequential inputs for the backbone LLM.}
    \label{fig:Model Architecture}
\end{figure*}

% \vspace{-9pt}
\subsection{Model Architecture}\label{AA}

% \guo{(Too many math notations and symbols in the section. Only need to keep the most essential steps with equations and there is no need to introduce a symbol for every simple step or variable.)}

Fig.~\ref{fig:Model Architecture} illustrates the overall model architecture. Given the static metadata \( X_{p} \) and wearable sensor data \( X_{s} \), we aim to provide specific health insights tailored to the task instruction \( X_{q} \). To achieve this, our framework includes four modules: a text embedder that encodes \( X_{p} \) and \( X_{q} \) into text token embeddings, a sensor encoder with an adapter to map \( X_{s} \) into sensor data embeddings, a backbone LLM to fuse and analyze all the provided information, and task-specific projection layers to produce predictions aligned with the downstream task. 

% \yifan{First paragraph: Text Embedder} 
\paragraph{Text Embedder} In order to instruct LLMs to understand the contextual information for wearable sensor data, we design a structured prompt template, which can provide rich personal static metadata information and a brief explanation of the task. As shown in Fig.~\ref{fig:Model Architecture}, each task-specific template follows a similar structure and consists of two components: static metadata \( X_{p}\) and task-specific instruction \( X_{q} \).
Both components are tokenized and embedded, producing a sequence of word embeddings, denoted as \( E_{p} \in \mathbb{R}^{L_{p} \times d} \) and \( E_{q} \in \mathbb{R}^{L_{q} \times d}\), where \( L_{p}\) and \( L_{q} \) are the lengths of token sequences and \( d \) is the dimension of word embeddings, respectively. We maintain consistency by adopting the same tokenizer and word embeddings from the backbone LLM. This ensures that \(d\) matches the hidden state dimension of the transformer blocks of the used LLM. 

% \yifan{Second paragraph: Multivariate Sensor Encoder with Adapter}
\paragraph{Multivariate Sensor Encoder with Adapter} For wearable sensor data, patching \cite{10.24963/ijcai.2024/895}  serves as an effective tokenization mechanism by dividing continuous sequences into discrete segments. 
% \yifan{ViT. As demonstraed SensorLM compared to other sensor encoder}
Building on this patch tokenization, we adapt a pre-trained Vision Transformer (ViT) \cite{dosovitskiy2020image} as the sensor encoder to leverage patch-level attention mechanisms for sensor data representation. Unlike traditional time series encoders such as PatchTST \cite{nie2022time}, the ViT jointly tokenizes both temporal and variable dimensions to enable cross-variable interactions within each patch.
Inspired by SensorLM \cite{zhang2025sensorlm}, which demonstrates the effectiveness of ViT for wearable sensor encoding, we leverage the ViT's transformer blocks to further model both local and long-range temporal dependencies across patches.
Also, the interactions among wearable data allow ViT to exploit proxy signals under weak supervision to identify specific glycemic events.

Mathematically, denote a sample of wearable data as 
% \( X_s = [x_{t,i}] \in \mathbb{R}^{T \times V} \), 
\( X_s \in \mathbb{R}^{T \times V} \), 
where 
% \(x_{t,i}\) denotes the aggregated (e.g., mean) value of sensor variate \(i\) within the \(t\)-th time interval, 
\(T\) is the total number of time intervals, and $V$ is the total number of sensor variables. Using a patch size of \( p_{t} \times p_{v} \), where 
\( p_{t} \) and \( p_{v} \) denote the temporal and variable patch sizes, respectively, the resulting number of patches is \( L_{s} = \frac{T}{p_{t}} \times \frac{V}{p_{v}} \). Finally, the ViT encoder transforms sensor data into a sequence of \( L_{s} \) embeddings, denoted by \(E_{s} \in \mathbb{R}^{L_{s} \times m}\), where \(m\) is the dimension of the original sensor data embeddings. These embeddings serve as input to the LLM. 

% As the backbone LLM has a different hidden dimension space, we need to align the sensor data embeddings with word embeddings. Following the insights from previous work \cite{zhang2025sensorlm}, we adopt a multi-layer perceptron (MLP) as the adapter \( \mathcal{P} (\,\cdot\,) \). Then, we obtain the final sensor data embeddings as:
% \begin{equation}
%   E_s \;=\; \mathcal{P}\!\bigl(e_s\bigr)
%   \label{eq:sensor_proj},
% \end{equation} where \( E_{s} \in \mathbb{R}^{L_{s} \times d} \).
% \yifan{Adapater parameters}

As the backbone LLM has a different hidden dimension space, we align the sensor embeddings with the LLM's latent space using a multi-layer perceptron (MLP) adapter. Specifically, the adapter $\mathcal{P}_{\theta_1}(\cdot)$ is a non-linear projection layer parameterized by $\theta_1$, which maps the sensor features into the joint embedding space:
\( \tilde{E}_s = \mathcal{P}_{\theta_1}(E_s) \in \mathbb{R}^{L_s \times d} \),
% \begin{equation}
%     E_s = \mathcal{P}_{\theta_1}(e_s) 
%     \in \mathbb{R}^{L_s \times d}\label{eq:sensor_proj},
%     \end{equation}
where $\theta_1$ denotes the trainable parameters of the adapter. This ensures that the dimensionality of the sensor embeddings matches the hidden state dimension $d$ of the backbone LLM.

% \yifan{Third paragraph: LLM and LoRA Fine-tuning}
\paragraph{LLM and LoRA Fine-tuning}
% \yifan{LLM auto-regression}
LLMs are autoregressive sequence models that learn contextualized representations by modeling dependencies over input token embeddings.
In our framework, we leverage this capability by constructing a unified embedding sequence that integrates static metadata, wearable sensor representations, and task-specific instructions.
Specifically, we concatenate the three embeddings \( E_{p}\), \( E_{q} \), and \( \tilde{E}_{s} \) to form the input into the pre-trained LLM backbone as \( E = \mathrm{concat} \bigl(E_p,\,\tilde{E}_s,\,E_q\bigr)\;\in\;\mathbb{R}^{L \times d} \),
% \begin{equation}\label{eq:concat_emb}
% E = \mathrm{concat} \bigl(E_p,\,E_s,\,E_q\bigr)\;\in\;\mathbb{R}^{L \times d},
% \end{equation} 
where \( L = L_p + L_s + L_q\). 

To capture the complex cross-modal interactions within this unified sequence, each transformer layer in the backbone performs multi-head attention (MHA).
Specifically, for each head $k \in \{1, \dots, H\}$, we define the query, key, and value matrices as:
\begin{equation}
Q_k = E W_k^Q, \quad K_k = E W_k^K, \quad V_k = E W_k^V \label{eq:qkv_head},
\end{equation}
where $W_k^Q, W_k^K, W_k^V \in \mathbb{R}^{d \times d_h}$ are the projection matrices and $d_h = d/H$ denotes the dimension of each attention head. Then, the operation to capture dependencies across metadata and sensor patches in each attention head is defined as:
\begin{equation}
Z_k = \mathrm{Attention}(Q_k, K_k, V_k) = \mathrm{Softmax}\left(\frac{Q_k K_k^\top}{\sqrt{d_h}}\right) V_k \label{eq:attn_head}.
\end{equation}
By aggregating each \( Z_k \in \mathbb{R}^{L \times d_h} \), we obtain \( Z \in \mathbb{R}^{L \times d}\) in each multi-head attention layer.

To efficiently adapt the LLM to downstream tasks, we adopt Low-Rank Adaptation (LoRA) \cite{hu2022lora,yeh2023navigating}, a parameter-efficient fine-tuning method that has been extended to various settings including resource-constrained edge devices \cite{11044641, yang2026fedkrso}.
% which enables parameter-efficient fine-tuning with fewer parameters. 
Specifically, during training, the pre-trained weight matrix \( W_0 \in \mathbb{R}^{a \times b}\) is frozen, and only two low-rank trainable matrices \( A \in \mathbb{R}^{r \times b} \) and \( B \in \mathbb{R}^{a \times r}\) are updated, where \( r \ll \min(a, b) \). For a linear projection $u = W_0 x$, the modified forward pass yields:
\begin{equation}
u = W_0 x + B A x.
\end{equation}

Specifically, as illustrated in Fig.~\ref{fig:Model Architecture}, we apply this adaptation to the weight matrices within both the MHA and Feed-Forward Network (FFN) modules across all $N$ transformer blocks. In the MHA, we target the query, key, and value projection matrices ($W^Q, W^K, W^V$), while in the FFN, the adaptation is applied to the two linear transformation matrices ($W_{in}, W_{out}$).

The collective set of these trainable low-rank parameters is denoted as $\theta_2$. 
Thus, the fine-tuning process of the backbone LLM can be represented as:
\( E_{\text{final}} = \mathcal{M}_{\theta_2}(E), \)
where $\mathcal{M}$ represents the transformer blocks of the backbone LLM.

Compared to full fine-tuning, LoRA can significantly reduce computational cost and memory requirements while also maintaining the representation capability of the LLM to adapt to new tasks.

% \yifan{Task-Specific Projection Layer}
\paragraph{Task-Specific Projection Layer}
% \yifan{cite MLP related work}
The final output of the backbone LLM is a sequence of hidden states $E_{\text{final}} \in \mathbb{R}^{L \times d}$, which are flattened into a single vector representation.
% $\bar{E} \in \mathbb{R}^{L \cdot d}$.
To adapt to diverse task objectives, we employ a projection head 
$\mathcal{G}_{\theta_3}(\cdot)$ parameterized by $\theta_3$, 
which maps the aggregated representation to the final prediction: 
\( \hat{Y} = \mathcal{G}_{\theta_3}(\text{Flatten} (E_{\text{final}})) \),
% $\hat{Y}$:
% \begin{equation}
% \hat{Y} = \mathcal{G}_{\theta_3}(\bar{E}),
% \end{equation}
where $\theta_3$ denotes the trainable parameters of the output layer.
For glucose forecasting, the vector is passed through a regression head to predict continuous glucose values. As for diabetes categorization, we use a linear classifier as the projection layer to get the final output, yielding multi-class diabetes stages. 
% The versatile design of this task-specific output layer allows our model to be easily adapted to specific diabetes-related tasks. 

By integrating the static metadata and sensor data through the fine-tuned model, the overall forward pass can be formulated as:
\begin{equation}
\hat{Y} = \mathcal{G}_{\theta_3} \left( \text{Flatten} \left( \mathcal{M}_{\theta_2} \left( \text{concat}(E_p, \mathcal{P}_{\theta_1}(E_s), E_q) \right) \right) \right),
\end{equation}
where $\theta_1, \theta_2, \text{ and } \theta_3$ represent the trainable parameters of the adapter, the LoRA module, and the task-specific head, respectively.

% \yifan{Training Objective Loss formulation}
To optimize the parameters $\Theta = \{\theta_1, \theta_2, \theta_3\}$ for varied downstream tasks, we define a generalized training objective.
The loss function is formulated as follows:
% \begin{equation}
% $$\mathcal{L}(\Theta) = 
%     \mathbb{E}_{(X_p, X_s, X_q, Y) \sim \mathcal{D}} \left[ \ell(\hat{Y}(\Theta), Y) \right]$$
% \end{equation}  
\begin{equation}
\mathcal{L}(\Theta) = 
    \mathbb{E}_{(X_p, X_s, X_q, Y) \sim \mathcal{D}} \left[ \ell(\hat{Y}(\Theta), Y) \right]
\end{equation}
where $\mathcal{D}$ represents the training dataset, and $\ell(\cdot)$ denotes the task-specific loss function. Here, $\hat{Y}(\Theta)$ represents the model's predicted output, and $Y$ denotes the corresponding ground-truth label.

\begin{figure*}[htbp]
    \centering
    \includegraphics[width=0.85\linewidth]{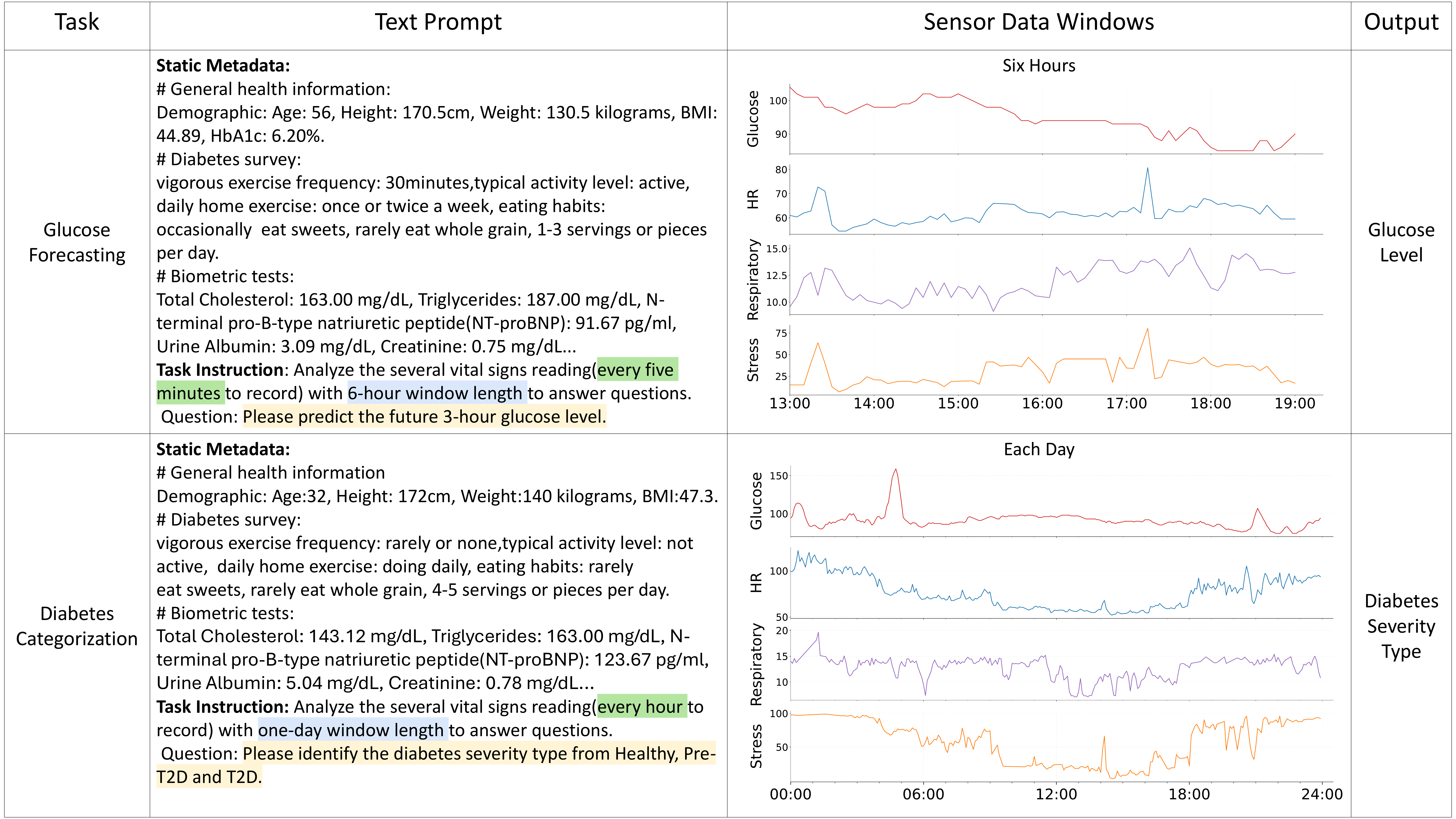}
    \caption{Examples of text prompts and sensor data used in two tasks.}
    \label{fig:examples of instructions}
    % \vspace*{-10pt}
\end{figure*}

\vspace{-4pt}
\subsection{Static Metadata Design}
% \yifan{Context information, cite Health-related work.We select follwing metadata in our method.} \yifan{This section introduces the data curation about personalization prompt across two tasks.}
Recent studies in health applications have demonstrated that incorporating specific static health factors and personalization can improve clinical decision support % \cite{Hall2018, Pasquel2021, Wang2024}, 
\cite{Pasquel2021, Wang2024}, 
which can provide critical contextual information of individual-level profiles. 
Building on this paradigm, we consider these static factors as essential textual context that guides the interpretation of dynamic wearable signals.

To unify text data and sensor data during training, we adopt a contextual prompt formulation, including personal static metadata information, the corresponding sensor data and task instructions, as shown in Fig.~\ref{fig:examples of instructions}. The templates of \( X_{p}\) and \( X_{q} \) are constructed as follows. First, the static metadata \( X_{p}\) provides the related health condition screening results that can be combined with the given wearable sensor data. The structured metadata components are organized as follows:
\begin{itemize}
\item \textbf{General Health Information (GHI)} 
assesses overall demographic contexts such as age, BMI, and HbA1c.

\item \textbf{Diabetes Survey (DS)} summarizes participant-specific lifestyle and dietary factors.
% summarizes participant-specific lifestyle and dietary factors

\item \textbf{Biometric Tests (Bio)} encompass a wide range of physical characteristics and biological tests performed on samples of blood, urine, or other tissues.
\end{itemize}

Second, the task instruction \( X_{q} \) is a task query to instruct LLMs to understand the text information and analyze multivariate wearable sensor data for specific predictive tasks. It is organized as: ``\textit{Analyze the several vital signs readings (\{\(F\)\} to record) with \{\(D\)\} to answer questions. Question: \{\(Q\)\}.}'' Here \{F\} indicates the sampling interval, \{D\} presents the time window length, and \{Q\} denotes the task specific question.

\vspace{-4pt}

\begin{table}[htbp]
\centering
\caption{Vital sign variables.}
\label{tab:vital_signs}
\small
\begin{tabularx}{\linewidth}{p{2.1cm} p{2.0cm} X}
\toprule
\textbf{Vital Sign} & \textbf{Sampling Rate} & \textbf{Description} \\
\midrule
Glucose Level 
& Five-minute 
& Interstitial glucose concentration (mg/dL) \\

Heart Rate 
& Minute-level 
& Average beats per minute (BPM) \\

Respiratory Rate 
& Minute-level 
& Breaths per minute derived from HRV and motion sensors \\

Stress Level 
& Minute-level 
& HRV-derived allostatic stress index (0--100) \\
\bottomrule
\end{tabularx}
\end{table}
\vspace{-8pt}

\begin{table}[htbp]
\centering
\caption{Data splitting statistics across diabetes groups.}
\label{tab:aireadi_summary}
\small
\setlength{\tabcolsep}{6pt}
\begin{tabular}{l c c c c}
\toprule
\textbf{Split} 
& \textbf{Healthy} 
& \textbf{Pre-T2D} 
& \multicolumn{2}{c}{\textbf{T2D}} \\
\cmidrule(lr){4-5}
& & & \textbf{Non-Insulin} & \textbf{Insulin} \\
\midrule
Train        & 253 & 142 & 195 & 50 \\
Validation   & 32  & 33  & 32  & 26 \\
Test         & 37  & 32  & 31  & 32 \\
\midrule
Total        & 322 & 207 & 258 & 108 \\
\bottomrule
\end{tabular}
\end{table}

\begin{table*}[t]
\centering
\caption{Participant characteristics across diabetes groups (mean(SD)).}
\label{tab:clinical_summary}
\footnotesize
\setlength{\tabcolsep}{6pt}
\begin{tabular}{lcccc}
\toprule
\textbf{Variable} 
& \textbf{Healthy} 
& \textbf{Pre-T2D} 
& \multicolumn{2}{c}{\textbf{T2D}} \\
\cmidrule(lr){4-5}
& & & \textbf{Non-Insulin} & \textbf{Insulin} \\
\midrule

Age (years)
& 59.07 (11.33)
& 59.64 (10.82)
& 60.91 (11.18)
& 60.21 (11.28) \\

BMI (kg/m$^2$)
& 28.49 (7.01)
& 30.19 (7.53)
& 32.08 (8.56)
& 34.30 (9.00) \\

Hemoglobin A1c (\%)
& 5.49 (0.70)
& 5.80 (0.82)
& 6.52 (1.33)
& 7.14 (1.72) \\

Total Cholesterol (mg/dL)
& 186.61 (40.21)
& 180.50 (40.97)
& 153.84 (40.56)
& 146.56 (42.27) \\

Triglycerides (mg/dL)
& 138.80 (80.03)
& 150.23 (144.13)
& 157.84 (101.10)
& 185.35 (172.33) \\

NT-proBNP (pg/mL)
& 93.97 (173.30)
& 123.34 (599.24)
& 119.21 (302.31)
& 185.64 (454.10) \\

Urine Albumin (mg/mL)
& 1.33 (5.63)
& 1.36 (3.81)
& 5.02 (20.71)
& 11.34 (55.65) \\

Creatinine (mg/mL)
& 0.86 (0.23)
& 0.87 (0.36)
& 0.92 (0.31)
& 1.22 (0.67) \\

\bottomrule
\end{tabular}
\end{table*}
\vspace{-3pt}

\section{Experiments}

% \yifan{Data, Model, Inmpletation Training Details}
\vspace{-4pt}
\subsection{Data}
\subsubsection{Dataset} 
% \yifan{Firsr paragraph: AI-READI dataset description.}
The dataset we used for our experiments is the controlled access AI-READI v2.0.0\footnote{\url{https://fairhub.io/datasets/2}} (Artificial Intelligence Ready and Equitable Atlas  for Diabetes Insights) Flagship Dataset \cite{aireadi_nature}. A total of 1,067 participants were recruited from three U.S. sites. 
The data collection protocol began with a one-hour pre-visit involving self-reported surveys regarding demographics and health history.
The 3-to-4-hour on-site visit comprised comprehensive clinical assessments, such as blood biomarkers and physical measurements. 
Subsequently, participants completed a 10-day post-visit period involving continuous monitoring at home. 
For our study, we extracted CGM data and wearable activity sensor data including heart rate (HR), respiratory rate (RR) and garmin stress level, as shown in Table \ref{tab:vital_signs}. All these recordings are resampled at five-minute intervals.
% Table \ref{tab:aireadi_summary} provides an overview of the dataset partitioning criteria across four groups. 

\subsubsection{Data Curation} The details of preprocessing are described below. 
% \yifan{Data, Task, Evalutaion metrics}

Resampling.
Based on glucose recordings, all vital sign data were resampled to a uniform frequency by calculating the mean values over every five-minute interval, to ensure consistency across time. After that, for each individual's resampled data, we removed incomplete recordings from the first and last day of data collection. 
% % \noindent\bf{Historical Window Design.} 
% Historical Window Design.
% Subsequently, we extracted segments of length W (historical window), from the long-term recordings. For glucose forecasting and diabetes categorization, we set W to 6 hours and 24 hours, respectively. 

% \noindent\textbf{Data Cleaning and Missing Data Imputation.} 
Data cleaning and missing data imputation.
For each window, we filter out cases where any vital sign has continuous missing data exceeding 1 hour.
Furthermore, we exclude any window where the overall ratio of missing data surpasses 5\%.
Finally, linear interpolation with bidirectional filling was employed to impute these remaining missing data points.

A total of 895 participants are retained for the final analysis, comprising 322 healthy individuals, 207 with pre-T2D, 258 individuals with T2D on oral medication, and 108 with T2D on insulin. As shown in Table \ref{tab:aireadi_summary}, we split AI-READI dataset, individual-wise, into training, validation, and test subsets.
Table \ref{tab:clinical_summary} shows general health information and biometric characteristics.
\vspace{8pt}

\subsubsection{Task Evaluation}
% \yifan{task and Metrics} 

\begin{table}[!ht]
\centering
\caption{Interpretable glucose (iGlu) measures.}
\label{tab:iglu_metrics}
\begin{tabular}{p{2.8cm} p{5.0cm}}
\toprule
\textbf{iGlu Parameters} & \textbf{Definition} \\
\midrule
Mean (mg/dL) & Average glucose level \\

CV (\%) & Coefficient of variation, defined as SD / mean \\

GMI & Glucose Management Indicator computed from mean CGM \\

TBR54 (\%) & Percent time spent below 54 mg/dL \\

TBR70 (\%) & Percent time spent below 70 mg/dL \\

TIR (\%) & Percent time spent between 70 and 180 mg/dL \\

TAR180 (\%) & Percent time spent above 180 mg/dL \\

TAR250 (\%) & Percent time spent above 250 mg/dL \\

\bottomrule
\end{tabular}
\end{table}
\vspace{-8pt} 

\paragraph{Glucose Forecasting} 
We first consider the task of glucose forecasting, which aims to predict future glucose trajectories to enable early detection of abnormal glucose levels. Following previous work \cite{pmlr-v259-gu25a, Zhu2025}, we set the historical window $D$ as 6 hours of sensor data at five-minute resolution, and the prediction horizon (PH) as 3 hours. Clinically, a 6-hour window captures typical postprandial glucose dynamics and the 3-hour horizon allows actionable lead time for intervention.

We evaluate the model performance using general regression metrics including Mean Absolute Error (MAE) and Root Mean Squared Error (RMSE), which have been widely used in benchmarking non-clinical time series data to quantify the deviation between predicted values and the ground truth. Also, to validate the clinical relevance of the predicted glucose levels, we further evaluate the methods in terms of glucose analysis using CGM-based iGlu measures \cite{Lutsker2026}.
Specifically, for each example, we compute a set of interpretable iGlu measures, including mean glucose, glucose variability, glycemic management indicator (GMI), and time-in-range related metrics, as shown in Table \ref{tab:iglu_metrics}. 

For each iGlu measure, we calculate the MAE across all test samples. 
To obtain a single summary score, these errors are averaged to form the iGlu Composite Error (iGlu-CE), where lower values indicate better overall agreement.
% Let $\mathbf{y}_i \in \mathbb{R}^{T}$ denote the observed future glucose sequence for sample $i$, and $\hat{\mathbf{y}}_i \in \mathbb{R}^{T}$ the predicted sequence, where $i \in \{1, \dots, M\}$ and $M$ is the total number of test samples.
Let $y_i \in \mathbb{R}^{T}$ denote the observed future glucose sequence for sample $i$, and $\hat{y}_i \in \mathbb{R}^{T}$ the predicted sequence, where $i \in \{1, \dots, M\}$ and $M$ is the total number of test samples.
Let $\mathcal{J}=\{1,\dots,J\}$ be the set of iGlu measures.
For each measure $j \in \mathcal{J}$, we define a function $g_j(\cdot)$ that maps a glucose sequence to a scalar score:
\vspace{-5pt}
% \begin{equation}
% s_{i,j} = g_j(\mathbf{y}_i), 
% \quad
% \hat{s}_{i,j} = g_j(\hat{\mathbf{y}}_i).
% \end{equation}
\begin{equation}
s_{i,j} = g_j(y_i), 
\quad
\hat{s}_{i,j} = g_j(\hat{y}_i).
\end{equation}

Finally, we aggregate the errors across all iGlu measures into a single iGlu-CE, defined as
\begin{equation}
\mathrm{iGluCE}
= \frac{1}{J} \sum_{j=1}^{J}
\left(
\frac{1}{M} \sum_{i=1}^{M}
\left| s_{i,j} - \hat{s}_{i,j} \right|\right).
\end{equation}

\paragraph{Diabetes Categorization}
% \yifan{This section focus on diabetes severity classification results analysis.} \guo{(There should be only three classes as mentioned by Dr. Shi).} 
Next, we consider another critical downstream task, diabetes categorization, as diabetes status significantly correlates with metabolic patterns and glucose regulation mechanisms. Consistent with previous studies % \cite{Lu2025, Fraser2025, henriques2025sweetdeep}, 
\cite{Lu2025, Fraser2025},
this task aims to distinguish individuals with healthy glucose regulation, pre-T2D, and established T2D (Non-Insulin and Insulin), 
% which is critical for early preventive intervention. 
which is critical for early preventive interventions before the onset of disease, as well as treatment monitoring and optimization in those already diagnosed with T2D.
Notably, diabetes subtype represents a relatively stable risk predictor, remaining constant over extended periods of observation. Therefore, we set the monitoring window $D$ as 24 hours for analysis. 

For diabetes categorization evaluation, we select the following metrics: (1) Accuracy and AUROC. Accuracy measures the proportion of correctly classified samples across all diabetes classes. In addition, AUROC is calculated as the area under the receiver operating characteristic (ROC) curve, which plots the True Positive Rate (TPR) against the False Positive Rate (FPR) at various threshold settings. (2) Sensitivity and Specificity. Sensitivity measures the proportion of actual positive cases correctly identified:
\begin{equation}
\text{Sensitivity}
=
\frac{\text{True Positives}}
{\text{True Positives} + \text{False Negatives}}
\end{equation}
Specificity measures the proportion of actual negative cases
correctly identified:
\begin{equation}
\text{Specificity}
=
\frac{\text{True Negatives}}
{\text{True Negatives} + \text{False Positives}}
\end{equation}

\begin{table*}[t]
\centering
\begin{threeparttable}
    \caption{Results of glucose level forecasting (PH = 3 hours) measured via average MAE and RMSE across five random seeds. Bold indicates the best result, and underline indicates the second best result in each group. ``Avg'' denotes the average results across four groups.}
    \label{tab:glucose_results}
    \footnotesize
    \setlength{\tabcolsep}{4pt} 
    \renewcommand{\arraystretch}{1.2}
    
    \begin{tabular}{l l cc cc cc cc cc}
    \toprule

    \multirow{2}{*}{\textbf{Setting}} & 
    \multirow{2}{*}{\textbf{Methods}} & 
    \multicolumn{2}{c}{\textbf{Healthy}} & 
    \multicolumn{2}{c}{\textbf{Pre-T2D}} & 
    \multicolumn{2}{c}{\textbf{Non-Insulin T2D}} & 
    \multicolumn{2}{c}{\textbf{Insulin T2D}} & 
    \multicolumn{2}{c}{\textbf{Avg}} \\

    \cmidrule(lr){3-4} \cmidrule(lr){5-6} \cmidrule(lr){7-8} 
    \cmidrule(lr){9-10} \cmidrule(lr){11-12}

    & & MAE & RMSE & MAE & RMSE & MAE & RMSE & MAE & RMSE & MAE & RMSE \\
    \midrule

    \multirow{3}{*}{\shortstack[l]{Transformer\\-based}}
    & PatchTST 
        & \res{24.28}{1.20} & \res{29.17}{2.35} 
        & \res{25.29}{1.53} & \res{30.37}{1.76} 
        & \res{29.03}{2.12} & \res{33.17}{2.20} 
        & \res{36.18}{2.01} & \res{39.62}{2.55} 
        & \res{28.70}{1.72} & \res{33.08}{2.22} \\

    & Crossformer 
        & \res{22.35}{0.92} & \secres{24.12}{1.88} 
        & \res{24.28}{1.45} & \res{29.32}{1.73} 
        & \res{28.41}{1.81} & \res{35.62}{1.94} 
        & \res{38.72}{2.88} & \res{40.11}{1.23} 
        & \res{28.44}{1.77} & \res{32.29}{1.70} \\

    & iTransformer 
        & \res{23.75}{1.14} & \res{26.19}{1.93} 
        & \res{27.07}{1.66} & \res{32.17}{2.43} 
        & \res{30.53}{1.85} & \res{34.22}{1.11} 
        & \res{37.82}{2.52} & \res{39.71}{2.39} 
        & \res{29.79}{1.79} & \res{33.07}{1.97} \\
    \midrule

    \multirow{3}{*}{Zero-shot}
    & Gemma-2 (2B) 
        & \res{24.18}{1.24} & \res{30.17}{2.52} 
        & \res{27.29}{1.31} & \res{32.04}{2.48} 
        & \res{31.18}{3.25} & \res{36.20}{3.37} 
        & \res{39.86}{2.91} & \res{41.66}{2.60} 
        & \res{30.63}{2.18} & \res{35.02}{2.74} \\

    & Mistral (7B) 
        & \res{23.17}{1.18} & \res{28.55}{2.11} 
        & \res{24.18}{1.59} & \res{34.13}{2.73} 
        & \res{31.18}{1.98} & \res{35.39}{2.02} 
        & \res{40.11}{2.14} & \res{42.29}{2.78} 
        & \res{29.66}{1.72} & \res{35.09}{2.41} \\

    & Llama3-Med42 (8B)
        & \res{26.15}{1.42} & \res{30.62}{2.45} 
        & \res{28.14}{1.83} & \res{35.18}{2.89} 
        & \res{30.21}{1.75} & \res{34.04}{2.82} 
        & \res{40.02}{2.26} & \res{42.12}{2.81} 
        & \res{28.63}{1.82} & \res{35.49}{2.74} \\
    \midrule

    \multirow{3}{*}{Few-shot}
    & Gemma-2 (2B) 
        & \res{28.14}{1.85} & \res{39.26}{3.52} 
        & \res{29.17}{2.42} & \res{38.18}{3.21} 
        & \res{34.18}{2.54} & \res{39.17}{2.58} 
        & \res{41.18}{2.03} & \res{41.78}{2.92} 
        & \res{33.17}{2.21} & \res{39.60}{3.06} \\

    & Mistral (7B)
        & \res{31.42}{2.68} & \res{38.24}{3.19} 
        & \res{31.25}{2.62} & \res{37.26}{3.14} 
        & \res{36.12}{2.87} & \res{38.16}{2.45} 
        & \res{43.17}{2.31} & \res{42.11}{2.86} 
        & \res{35.49}{2.62} & \res{38.94}{2.91} \\

    & Llama3-Med42 (8B) 
        & \res{34.56}{3.36} & \res{39.18}{3.31} 
        & \res{35.17}{2.71} & \res{39.26}{3.33} 
        & \res{39.14}{2.24} & \res{40.17}{2.67} 
        & \res{44.16}{2.42} & \res{42.19}{2.01} 
        & \res{38.26}{2.68} & \res{40.20}{2.83} \\
        \midrule

    \multirow{3}{*}{GlyLLM}
    & Gemma-2 (2B) 
        & \res{22.71}{1.15} & \res{25.16}{1.84} 
        & \res{23.13}{1.48} & \res{27.62}{2.46} 
        & \res{25.18}{1.85} & \res{30.87}{1.69} 
        & \res{34.29}{2.95} & \res{38.29}{2.47} 
        & \res{26.33}{1.86} & \res{30.49}{2.12} \\

    & Mistral (7B)
        & \secres{20.12}{0.88} & \res{24.64}{1.76} 
        & \bestres{20.05}{1.15} & \secres{26.17}{2.25} 
        & \secres{24.18}{1.26} & \bestres{25.12}{1.15} 
        & \secres{35.18}{2.18} & \secres{37.59}{2.24} 
        & \secres{24.88}{1.37} & \secres{28.38}{1.85} \\

    & Llama3-Med42 (8B)
        & \bestres{19.50}{0.79} & \bestres{22.19}{1.55} 
        & \secres{20.14}{1.12} & \bestres{25.03}{2.18} 
        & \bestres{23.62}{1.31} & \secres{26.93}{1.22} 
        & \bestres{33.32}{2.76} & \bestres{37.35}{2.12} 
        & \bestres{24.15}{1.50} & \bestres{27.88}{1.77} \\
    \bottomrule
    \end{tabular}
\end{threeparttable}
\end{table*}

\subsection{Model Implementation Details}

\begin{figure}[!ht]
\centering

% ---------- Zero-shot ----------
\begin{subfigure}{\linewidth}
\centering
\begin{tcolorbox}[
  width=\linewidth,
  colback=blue!5,
  colframe=black,
  boxrule=0.8pt,
  arc=2pt,
  left=6pt,
  right=6pt,
  top=6pt,
  bottom=6pt
]
\footnotesize \itshape
You are a professional diabetes doctor in the emergency department. Your role is to analyze the following glucose data over a 6-hour window and predict the patient's future blood glucose levels for the next 3 hours.

\vspace{0.6em}
A [age]-year-old [sex] individual has an HbA1c of [HbA1c]\% and a BMI of [BMI]. 

\#\# Diabetes Survey: [Diabetes Survey]

\#\# Biometric tests: [Biometric tests]

Based on continuous glucose monitoring (CGM) data recorded every 5 minutes over the past 6 hours ([historical\_readings]), predict the patient's glucose levels for the next three hours at 5-minute intervals. 
\end{tcolorbox}
\vspace{-10pt}
\caption{Zero-shot prompt}
\end{subfigure}

\vspace{0.8em}

% ---------- Few-shot ----------
\begin{subfigure}{\linewidth}
\centering
\begin{tcolorbox}[
  width=\linewidth,
  colback=blue!5,
  colframe=black,
  boxrule=0.8pt,
  arc=2pt,
  left=6pt,
  right=6pt,
  top=6pt,
  bottom=6pt
]
\footnotesize \itshape
You are a professional diabetes doctor \ldots the next 3 hours.

\vspace{0.6em}
A [age]-year-old [sex] individual has an HbA1c of [HbA1c]\% and a BMI of [BMI]. 

\#\# Diabetes Survey: [Diabetes Survey]

\#\# Biometric tests: [Biometric tests]

\vspace{0.6em}
Here are 3 examples:

Example \#1 \\
Historical reading: [110.0, 112.5, 114.8, \ldots, 120.3] mg/dL \\
Future readings: [121.3, 117.8, 101.8, \ldots, 122.8] mg/dL

\vspace{0.4em}
Example \#2 \\
Historical reading: [91.6, 95.3, 203.3, \ldots, 108.1] mg/dL \\
Future readings: [110.4, 120.2, 110.2, \ldots, 102.8] mg/dL

\vspace{0.4em}
Example \#3 \\
Historical reading: [82.1, 92.5, 104.1, \ldots, 90.3] mg/dL \\
Future readings: [85.3, 82.4, 80.1, \ldots, 73.4] mg/dL

Based on continuous glucose monitoring (CGM) data \ldots at 5-minute intervals.
% Based on continuous glucose monitoring (CGM) data recorded every 5 minutes over the past 6 hours ([historical\_readings]), predict the patient's glucose levels for the next three hours at 5-minute intervals. 

\end{tcolorbox}
\vspace{-10pt}
\caption{Few-shot prompt}
\end{subfigure}
\vspace{-15pt}
\caption{Prompt templates used for glucose forecasting.}
\label{fig:prompt_templates}
\end{figure}
% \vspace{-4pt}

\subsubsection{Module}
\paragraph{Sensor encoder.} The sensor encoder that we choose is the pre-trained Google\_vit-base-patch16-224\footnote{\url{https://huggingface.co/google/vit-base-patch16-224}}. It is a transformer-based pre-trained model that processes images as sequences of patches, applying the transformer encoder directly to these patch embeddings. All input images are resized to 224$\times$224 pixels and divided into 16$\times$16 patches, resulting in a sequence of 196 patches. Each patch is linearly embedded into a 768-dimensional vector, with positional embeddings added to retain spatial information. 
% It outputs patch embeddings of 196 patches, which is input into the LLM as 196 tokens after the adapter module.

\paragraph{LLMs and LoRA Finetuning} We use the Llama3-Med42-8B 
\cite{christophe2024med42}
model, which is built on the Llama3 architecture and fine-tuned using specialized clinical data. For the ablation study, we also explore two other LLMs: 
Gemma-2-2B \cite{team2024gemma} and Mistral-7B-v0.1 \cite{jiang2023mistral7b}. 
These LLMs are available at Hugging Face\footnote{\url{https://huggingface.co/}}.
To efficiently fine-tune LLMs, we employ a LoRA module of rank \( r = 16 \) and \( \alpha = 32 \).

% \vspace{-5pt}
\subsubsection{Training Settings}
For both tasks, the training process aims to optimize the trainable parameters $\Theta = \{\theta_1, \theta_2, \theta_3\}$ by minimizing $\mathcal{L}(\Theta)$.
Specifically, for glucose forecasting, GlyLLM is trained and evaluated independently on each group for 10 epochs, using a batch size of 64. The parameters are optimized by minimizing the Mean Squared Error (MSE) loss between the predicted and ground-truth glucose levels.
In the diabetes categorization task, GlyLLM is trained across three target groups (Healthy, Pre-T2D, and T2D) together for 5 epochs, using a batch size of 128. In this setting, the model is optimized by minimizing the cross-entropy loss to achieve accurate multi-class classification. All experiments are conducted on 4 NVIDIA A6000 GPUs. 
We employ the Adam optimizer with a learning rate of $1 \times 10^{-4}$ to update $\Theta$. To enhance training efficiency and optimize memory usage, we utilize DeepSpeed ZeRO-2 % \cite{rajbhandari2020zero}
to accelerate the training process.
\vspace{-4pt}

\begin{table*}[t]
\centering
\caption{Results of glucose forecasting evaluated via average iGlu-CE across five random seeds. Bold indicates the best result, and underline indicates the second best result in each group. ``Avg'' denotes the average results across four groups}
\label{tab:iglu_results}

\footnotesize
\setlength{\tabcolsep}{6pt} 
\renewcommand{\arraystretch}{1.0}

\begin{tabular}{l l c c c c c}
\toprule
\textbf{Setting} & \textbf{Methods} 
& \textbf{Healthy} 
& \textbf{Pre-T2D} 
& \textbf{Non-Insulin T2D} % 
& \textbf{Insulin T2D} 
& \textbf{Avg} \\
\midrule

\multirow{3}{*}{\shortstack[l]{Transformer\\-based}}
& PatchTST      
    & \res{12.16}{0.71} & 
    \res{12.26}{0.78} & 
    \res{12.93}{0.72} & 
    \res{14.62}{0.75} & 
    \res{12.98}{0.74} \\

& Crossformer   
    & \res{10.66}{0.63} & 
    \secres{9.04}{0.58} & 
    \res{12.71}{0.79} & 
    \res{14.41}{0.68} & 
    \res{11.71}{0.67} \\

& iTransformer  
    & \res{11.02}{0.68} & 
    \res{11.14}{0.62} & 
    \res{11.27}{0.65} & 
    \res{13.81}{0.70} & 
    \res{11.81}{0.66} \\
\midrule

\multirow{3}{*}{Zero-shot}
& Gemma-2 (2B)  
    & \res{13.18}{0.81} & 
    \res{14.83}{0.92} & 
    \res{16.57}{0.96} & 
    \res{21.61}{1.12} & 
    \res{16.55}{0.95} \\

& Mistral (7B) 
    & \res{12.89}{0.78} & 
    \res{13.71}{0.88} & 
    \res{16.81}{0.84} & 
    \res{20.32}{1.05} & 
    \res{15.93}{0.89} \\

& Llama3-Med42 (8B)  & 
    \res{12.78}{0.74} & 
    \res{12.13}{0.75} & 
    \res{15.60}{0.78} & 
    \res{19.18}{0.95} & 
    \res{14.91}{0.81} \\
\midrule

\multirow{3}{*}{Few-shot}
& Gemma-2 (2B)  & 
    \res{13.82}{0.85} & 
    \res{14.03}{0.83} & 
    \res{17.79}{0.89} & 
    \res{22.73}{1.20} & 
    \res{17.09}{0.94} \\

& Mistral (7B) & 
    \res{13.12}{0.82} & 
    \res{13.82}{0.79} & 
    \res{17.25}{0.86} & 
    \res{22.11}{1.18} & 
    \res{16.58}{0.91} \\

& Llama3-Med42 (8B)  & 
    \res{12.16}{0.75} & 
    \res{13.67}{0.74} & 
    \res{16.18}{0.85} & 
    \res{20.12}{1.02} & 
    \res{15.53}{0.84} \\
\midrule

\multirow{3}{*}{GlyLLM}
& Gemma-2 (2B)  & 
    \res{7.83}{0.55}  & 
    \res{10.12}{0.62} & 
    \res{10.70}{0.64} & 
    \res{13.96}{0.72} & 
    \res{10.65}{0.63} \\
    
& Mistral (7B) & 
    \secres{6.63}{0.46} & 
    \res{9.17}{0.54}  & 
    \secres{10.67}{0.58} & 
    \secres{14.08}{0.75} & 
    \secres{10.14}{0.58} \\
    
& Llama3-Med42 (8B)  & 
    \bestres{6.08}{0.52} & 
    \bestres{8.14}{0.35} & 
    \bestres{10.31}{0.59} & 
    \bestres{13.04}{0.68} & 
    \bestres{9.39}{0.54} \\
\bottomrule
\end{tabular}
\end{table*}

\subsection{Baselines} 
% \yifan{Baselines}
% \yifan{This section focus on how we consider different traditional methods and deep learning methods for comprehensive comparison.}
% \guo{(You will also need to consider different LLM methods.)}
% \subsubsection{Baselines}
For comprehensive comparison, we first choose traditional ML-based methods. 
For glucose forecasting, the baseline methods encompass transformer-based models including PatchTST \cite{nie2022time}, Crossformer \cite{zhang2023crossformer} and iTransformer \cite{liu2024itransformer}. 
For diabetes categorization, we compare with two ML-based methods, including MLP \cite{Hornik1989} and LSTM \cite{Hochreiter1997}.

Additionally, to demonstrate the effectiveness of our proposed method compared to using LLMs directly, we select two state-of-the-art prompting techniques: 1) Zero-Shot Prompting \cite{kojima2022large}, 2) Few-Shot Prompting \cite{brown2020language}, as shown in Fig.~\ref{fig:prompt_templates} using glucose forecasting as an example.

\paragraph{Zero-Shot Prompting} In zero-shot prompting \cite{kojima2022large}, LLMs are required to perform a task based solely on a task description and a single input instance, without any examples. In this setting, we apply the baseline \cite{Zhu2025} structured template and construct the prompts that include demographic information alongside six hours of historical CGM readings.

\paragraph{Few-Shot Prompting}
LLMs exhibit in-context learning capabilities \cite{brown2020language}, allowing few-shot prompting to provide additional task guidance beyond zero-shot prompting. In few-shot prompting, LLMs are provided with a limited number of task-specific examples ``shots'', each paired with its corresponding target output. These examples provide contextual guidance that helps LLMs infer the nature and structure of the target task. In our setting, we provide 3-shot examples.
\vspace{-6pt}

\begin{table*}[t]
\centering
\caption{Results of diabetes categorization measured via average Macro-AUROC, Accuracy, Macro-Sensitivity, and Macro-Specificity across five random seeds. These metrics are computed using a one-vs-rest strategy and macro-averaged across classes. Bold indicates the best result, and underline indicates the second best result.}
\label{tab:screening_results}

\footnotesize
\setlength{\tabcolsep}{6pt} 
\renewcommand{\arraystretch}{1.2}

\begin{tabular}{l l c c c c}
\toprule
\textbf{Setting} & \textbf{Methods} 
& \textbf{Macro-AUROC} 
& \textbf{Accuracy} 
& \textbf{Macro-Sensitivity} 
& \textbf{Macro-Specificity} \\
\midrule

\multirow{2}{*}{ML-based}
& MLP   & \res{66.17}{2.25} & \res{60.23}{2.10} & \res{61.27}{1.15} & \res{64.82}{1.20} \\
& LSTM  & \res{69.16}{2.30} & \res{62.73}{3.18} & \res{64.72}{1.22} & \res{65.17}{1.19} \\
\midrule

\multirow{3}{*}{Zero-shot}
& Gemma-2 (2B)          & \res{67.78}{2.28} & \res{62.18}{2.15} & \res{65.17}{1.21} & \res{63.52}{1.18} \\
& Mistral (7B) & \res{71.17}{2.32} & \res{64.76}{2.20} & \res{65.07}{1.23} & \res{65.39}{1.21} \\
& Llama3-Med42 (8B)     & \res{73.61}{3.35} & \res{64.87}{2.22} & \res{66.37}{1.25} & \res{66.64}{1.24} \\
\midrule

\multirow{3}{*}{Few-shot}
& Gemma-2 (2B)          & \res{66.78}{2.40} & \res{57.75}{3.30} & \res{57.34}{1.35} & \res{61.27}{1.32} \\
& Mistral (7B) & \res{68.18}{2.42} & \res{60.12}{2.33} & \res{62.13}{1.38} & \res{62.57}{1.35} \\
& Llama3-Med42 (8B)     & \res{65.78}{3.38} & \res{58.05}{2.28} & \res{59.32}{1.30} & \res{61.03}{1.29} \\
\midrule

\multirow{3}{*}{GlyLLM}
& Gemma-2 (2B)          & \res{71.28}{1.88} & \res{64.71}{1.77} & \res{64.33}{0.94} & \res{66.23}{1.78} \\
& Mistral (7B) & \secres{75.67}{1.81} & \secres{65.67}{1.73} & \secres{66.81}{0.92} & \secres{67.51}{1.72} \\
& Llama3-Med42 (8B)     & \bestres{78.15}{1.74} & \bestres{66.27}{1.69} & \bestres{67.53}{0.83} & \bestres{71.06}{1.68} \\
\bottomrule
\end{tabular}
\end{table*}

% \subsection{Experimental Results \guo{(Experimental Results)}}

\section{Results Analysis}

\subsection{Glucose Forecasting}

\paragraph{\textbf{Glucose Forecasting Level Analysis}} First, we quantitatively compare our model with state-of-the-art baselines. As shown in Table \ref{tab:glucose_results}, our method reduces the RMSE by an average of 5.57\%, 12.11\%, and 13.66\% for Gemma-2-2B, Mistral-7B-v0.1, and Llama3-Med42-8B, compared to the best transformer-based methods. 
% (32.29 vs 30.49), (32.29 vs 28.38), (32.29 vs 27.88)
In some cases, our evaluation results reveal that none of these LLMs can produce clinically meaningful glucose predictions in the zero-shot setting. Instead of generating the requested numerical predictions, most LLMs' responses consist of explanations of glucose levels. Also, if we give a few examples (3-shot), LLMs mostly repeat the same glucose trend in future predictions. 
This observation shows that excessively long textual contexts can lead to unreliable LLM outputs, consistent with prior findings \cite{tan2024language}.
% \vspace{-11pt}

\paragraph{\textbf{CGM-based iGlu Composite Error (iGlu-CE) Evaluation}} 
Next, as shown in Table \ref{tab:iglu_results}, our method reduces the iGlu-CE by an average of 9.05\%, 13.41\%, and 19.81\% for Gemma-2-2B, Mistral-7B-v0.1, and Llama3-Med42-8B, respectively, compared to the best transformer-based methods. 
% (11.71 vs 10.65), (11.71 vs 10.14), (11.71 vs 9.39)
This finding indicates that GlyLLM can improve similarity between observed and generated CGM-based composite scores. Specifically, in two T2D groups (Non-Insulin and Insulin), the result shows that our method can reduce iGlu-CE by 8.52\% (11.27 vs 10.31) and 5.58\% (13.81 vs 13.04) for Llama3-Med42-8B. 
Overall, the improvement is more evident on iGlu-CE, suggesting that our framework can identify clinically relevant glycemic patterns beyond point-wise glucose prediction accuracy.

\subsection{Diabetes Categorization}

\paragraph{\textbf{Overall Classification Reliability Analysis}} As shown in Table \ref{tab:screening_results}, our method improves Macro-AUROC by 3.14\%, 9.49\%, and 13.08\% for Gemma-2-2B, Mistral-7B-v0.1, and Llama3-Med42-8B compared to the best ML-based methods.
% (69.11 vs 71.28), (69.11 vs 75.67), (69.11 vs 78.15)
Meanwhile, consistent gains in accuracy are also observed, with relative improvements of 3.16\%, 4.69\%, and 5.64\% across the three LLM backbones, indicating more reliable diabetes categorization performance.
% (62.73 vs 64.71), (62.73 vs 65.67), (62.73 vs 66.27)
Another observation is that in the zero-shot setting, the results reveal that LLMs can improve Macro-AUROC by 2.91\% (69.16 vs 71.17) and 6.43\% (69.16 vs 73.61) for Mistral-7B-v0.1 and Llama3-Med42-8B, respectively.
% (69.16 vs 71.17), (69.16 vs 73.61)
This suggests that, in multiple-choice formatted tasks, LLMs are capable of leveraging historical glucose trends to abstract semantic representations, which differs from glucose forecasting. Although the same prompting strategy is adopted for both tasks, glucose forecasting places substantially higher demands on semantic abstraction of long-horizon CGM patterns. 

\paragraph{\textbf{Class-wise Detection Evaluation}} In addition, we further analyze Macro-Sensitivity and Macro-Specificity to evaluate class-wise detection performance. As shown in Table \ref{tab:screening_results}, GlyLLM demonstrates consistent improvements in most cases, with Macro-Sensitivity increasing by up to 4.34\% (64.72 vs 67.53) and Macro-Specificity improving by as much as 9.04\% (65.17 vs 71.06), with the largest improvements observed on the Llama3-Med42-8B. 
From this perspective, GlyLLM can also identify healthy, pre-T2D, and T2D individuals, advancing more effective class-wise diagnosis in diabetes categorization.

\begin{table}[htbp]
\centering
\caption{Results of overall ablation study when removing sensor data and static metadata components across two tasks. Here, the results are averaged across four groups in glucose forecasting.
``w/o'' indicates the removal of metadata component from our model.}
\label{tab:ablation_summary}

\resizebox{\linewidth}{!}{%
    \footnotesize
    \setlength{\tabcolsep}{2pt} 
    \renewcommand{\arraystretch}{1.25} %
    \begin{tabular}{l l cc c} 
    \toprule
    \textbf{Setting} & \textbf{Method} &
    \multicolumn{2}{c}{\textbf{Glucose Forecasting}} &
    \textbf{Diabetes Categorization} \\
    \cmidrule(lr){3-4} \cmidrule(lr){5-5}
     &  & MAE & RMSE & AUROC \\
    \midrule
    
    \multirow{6}{*}{\shortstack[l]{Gemma-2\\(2B)}}
    & w/o sensor data        & \res{29.42}{1.25} & \res{34.71}{2.19} & \res{66.84}{1.45} \\
    & w/o static metadata    & \res{28.28}{1.18} & \res{33.81}{2.15} & \res{66.11}{1.40} \\
    & \hspace{0.5em} w/o GHI   & \res{28.18}{1.25} & \res{33.74}{2.21} & \res{67.07}{1.35} \\
    & \hspace{0.5em} w/o DS    & \res{28.16}{1.32} & \res{33.12}{2.15} & \res{66.71}{1.38} \\
    & \hspace{0.5em} w/o Bio   & \res{28.22}{1.24} & \res{33.63}{2.13} & \res{66.82}{1.36} \\
    & ours                 & \res{26.33}{1.86} & \res{31.49}{2.12} & \res{71.28}{1.88} \\
    \midrule
    
    \multirow{6}{*}{\shortstack[l]{Mistral\\(7B)}}
    & w/o sensor data        & \res{27.87}{1.80} & \res{31.82}{2.68} & \res{70.65}{1.30} \\
    & w/o static metadata    & \res{27.04}{1.65} & \res{31.13}{2.55} & \res{68.04}{1.28} \\
    & \hspace{0.5em} w/o GHI   & \res{26.81}{1.62} & \res{30.34}{2.48} & \res{68.71}{1.25} \\
    & \hspace{0.5em} w/o DS    & \res{26.37}{1.58} & \res{30.21}{2.45} & \res{69.53}{1.26} \\
    & \hspace{0.5em} w/o Bio   & \res{26.53}{1.60} & \res{30.43}{2.46} & \res{68.84}{1.27} \\
    & ours                 & \res{24.88}{1.37} & \res{28.38}{1.85} & \res{75.67}{1.81} \\
    \midrule
    
    \multirow{6}{*}{\shortstack[l]{Llama3-Med42\\(8B)}}
    & w/o sensor data        & \res{27.42}{1.65} & \res{31.86}{2.58} & \res{74.52}{1.15} \\
    & w/o static metadata    & \res{27.37}{1.55} & \res{31.17}{2.45} & \res{69.43}{1.12} \\
    & \hspace{0.5em} w/o GHI   & \res{26.81}{1.48} & \res{30.26}{2.38} & \res{70.48}{1.08} \\
    & \hspace{0.5em} w/o DS    & \res{26.10}{1.42} & \res{30.67}{2.40} & \res{71.54}{1.10} \\
    & \hspace{0.5em} w/o Bio   & \res{26.16}{1.45} & \res{30.21}{2.35} & \res{70.24}{1.11} \\
    & ours                 & \res{24.15}{1.50} & \res{27.88}{1.77} & \res{78.15}{1.74} \\
    \bottomrule
    \end{tabular}%
}
\end{table}

% \vspace{-10pt}
\subsection{Ablation Study} 
To examine the impact of static metadata information and sensor data in the proposed framework, we conduct ablation experiments by removing sensor data or static metadata components within the framework. The results depicted in Table \ref{tab:ablation_summary} show that removing either of them degrades model performance. Specifically, for glucose forecasting, it shows that removing sensor data results in RMSE increases of 10.23\% (31.49 vs 34.71), 12.12\% (28.38 vs 31.82), and 12.49\% (27.88 vs 31.86), suggesting that sensor data can provide rich information about behavior patterns to help identify future glucose trends. In diabetes categorization, removing static metadata leads to an AUROC decrease of 7.25\% (71.28 to 66.11), 10.08\% (75.67 to 68.04), and 11.16\% (78.15 to 69.43). Another observation is that removing DS and Bio metadata results in more severe model degradation compared to GHI. This suggests that LLMs are able to identify DS and Bio metadata as more critical priors at decision time for glycemic assessment.

% \subsection{Fine-tuning Efficiency Analysis}

% To evaluate the fine-tuning efficiency of GlyLLM, we considering different train ratio data to assess the performance changes. In detail, we design train ratio: \{0.1,0.2,0.4,0.8,1.0\} to analyze the model performance.

\vspace{-14pt}
\section{Conclusion}
% \yifan{This section summarizes our work contribution for glycemic assessment in T2D.}
In this paper, we propose GlyLLM, an LLM-powered framework for clinically personalized glycemic assessment in type 2 diabetes.
Our evaluation on diverse tasks on the AI-READI dataset shows that our proposed method outperforms baseline methods across two tasks. The proposed framework has the potential to advance diabetes care by providing insights to deliver more proactive clinical interventions in glycemic assessment. 
By integrating static metadata (GHI, DS, and Bio) and sensor data through LLMs, our framework can provide accurate glucose forecasting and effective diabetes categorization for personalized T2D care. In detail, we find that diabetes surveys and biometric tests have a more substantial impact on glycemic assessment compared to general health information, which reveals that diabetes-related factors are more important in diabetes care.
Importantly, we observe that prompting-based approaches for sensor data understanding exhibit limited practical utility and, in some cases, underperform traditional ML-based methods. These results suggest that directly instructing LLMs to interpret raw sensor data for semantic abstraction is insufficient, particularly when the input context spans long temporal sequences.

% \yifan{Second paragraph: Our work limitaions.} 
\textbf{Limitations.} This work highlights the potential of LLMs for glycemic assessment in T2D. However, its direct clinical applicability remains limited, as the approach has not yet been validated in real-world settings. Our experiments are confined to the AI-READI v2.0.0 dataset, and we have not assessed the model's performance on other real-world datasets or tasks. 
% Additionally, our current framework uses an ImageNet-pretrained ViT as the frozen sensor encoder. Exploring sensor-domain pretrained encoders or end-to-end fine-tuning of the encoder may further improve performance, which we leave for future work.

% \yifan{Third paragraph: Future work.} 
\textbf{Future Work.} 
To capture complex modality information, we use modality-specific encoders to extract the representative semantic information and fuse them through LLMs. A promising direction for future work is to explore sensor-domain pre-trained encoders and time-series foundation models for more comprehensive temporal dynamics modeling. 
In addition, we aim to integrate additional data types such as ECG signals and retinal images within the AI-READI dataset to explore whether more complex data fusion can provide deep insights into T2D care.
\vspace{-10pt}

\section*{Acknowledgment}
Y. Gao, Y. Shi, and Y. Guo were partially supported by a seed grant from UT San Antonio Office of Research and Innovation and NSF Grant CMMI-2222670. Y. Gong was partially supported by NSF Grant CNS-2611068. 
%We also appreciate the anonymous reviewers for their insightful comments.

% \section*{References}

\bibliographystyle{yifan}
\bibliography{yifan}

% \appendix

% \subsection{Dataset Preparation}\label{apd:first} \yifan{delete}
% \subsubsection{AI-READI}\label{apd:aireadi}
% AI-READI \cite{ai-readi_dataset2024} is a dataset consisting of data collected from individuals with and without T2D and harmonized across 3 data collection sites. The dataset is developed under the NIH Bridge2AI Program to support artificial  intelligence and machine learning on T2D. The version 2.0.0 of AI-READI consists of 1067 participants who are categorized into four groups based on their diabetes status: healthy, pre-diabetes, T2D on oral medication and T2D on insulin. This dataset is comprehensive in terms of demographics and health conditions. The important structure of this dataset is its multimodal data types, where participants were monitored over ten days using different wearable devices, such as Dexcom G6 CGM for real-time blood glucose, a Garmin Vivosmart 5 for physical activity and heart rate variability, and various questionnaires screening. Table \ref{tab:vital_signs} outlines the vital signs we select to analysis. Table \ref{tab:aireadi_summary} provides an overview of the dataset partitioning strategy and participant characteristics.
% \vspace{12pt}
% \color{red}

\end{document}